\documentclass{article}

\usepackage{arxiv}

\usepackage[utf8]{inputenc} % allow utf-8 input
\usepackage[T1]{fontenc}    % use 8-bit T1 fonts
\usepackage{hyperref}       % hyperlinks
\usepackage{url}            % simple URL typesetting
\usepackage{booktabs}       % professional-quality tables
\usepackage{amsfonts}       % blackboard math symbols
\usepackage{nicefrac}       % compact symbols for 1/2, etc.
\usepackage{microtype}      % microtypography
\usepackage{lipsum}
\usepackage{graphicx}
\usepackage{nicematrix}
\usepackage{colortbl}
\usepackage{float}

\graphicspath{ {./images/} }

\title{Multimodal Multi-Hop Question Answering Through a Conversation Between Tools and Efficiently Finetuned Large Language Models}

\author{
 Hossein Rajabzadeh \\
  University of Waterloo \\
  \texttt{hossein.rajabzadeh@uwaterloo.ca} \\
   \And
 Suyuchen Wang \\
  Mila\\ 
  DIRO, Université de Montréal \\
  \texttt{suyuchen.wang@umontreal.ca} \\
    \And
  Hyock Ju Kwon \\
  University of Waterloo \\
  \texttt{hjkwon@uwaterloo.ca} \\
    \And
 Bang Liu \\
  Mila\\
  Université de Montréal \\
  \texttt{bang.liu@umontreal.ca} \\
}

\begin{document}
\maketitle
\begin{abstract}
We employ a tool-interacting divide-and-conquer strategy enabling large language models (LLMs) to answer complex multimodal multi-hop questions. In particular, we harness the power of large language models to divide a given multimodal multi-hop question into unimodal single-hop sub-questions to be  answered by the appropriate tool from a predefined set of tools. After all  corresponding tools provide the LLM with their answers, the LLM generates the next relevant unimodal single-hop question. To increase the reasoning ability of LLMs, we prompt chatGPT to generate a tool-interacting divide-and-conquer dataset. This dataset is then used to efficiently finetune the corresponding LLM.
To assess the effectiveness of this approach, we conduct an evaluation on two recently introduced complex question-answering datasets. The experimental analysis demonstrate substantial improvements over existing state-of-the-art solutions, indicating the efficacy and generality of our strategy \footnote{Codes and data will be released soon.}.
\end{abstract}

% keywords can be removed
%\keywords{First keyword \and Second keyword \and More}

\section{Introduction}
Large language models (LLMs), such as ChatGPT \cite{radford2019language}, have demonstrated conspicuous reasoning abilities in few-shot settings across various tasks. However, answering multi-hop questions, which require sequential and compositional reasoning, remains a formidable challenge for LLMs \cite{brown2020language, bommasani2021opportunities, wei2022chain}. This challenge exposes serious limitations of LLMs, such as untruthfulness, false reasoning, and hallucination \cite{bang2023multitask, alkaissi2023artificial, press2022measuring, azamfirei2023large, kasneci2023chatgpt, ferrara2023should, creswell2022faithful}.

% Breaking the input question into a reasoning chain for LLMs has been widely explored by Self-Ask \cite{press2022measuring}, Least-to-most prompting \cite{zhou2022least}, ReAct \cite{yao2022react}, and \cite{wei2022chain}. While these techniques are effective for single-hop questions, it is still challenging for the case of multimodal multi-hop (MMH) reasoning questions, in which the subsequent questions are conditional to the answers of the precedent sub-questions. 

To enable LLMs to break down the input question into a reasoning chain, several techniques have been proposed, such as Self-Ask \cite{press2022measuring}, Least-to-most prompting \cite{zhou2022least}, ReAct \cite{yao2022react}, mm-ReAct \cite{yang2023mm}, Vipergpt \cite{suris2023vipergpt}, Chameleon \cite{lu2023chameleon}, ART \cite{paranjape2023art}, and CoT\cite{wei2022chain}. These methods work well for single-hop questions, but they face difficulties when dealing with multimodal multi-hop (MMH) reasoning questions, where the answers of the previous sub-questions affect the formulation of the subsequent ones. Besides, ReAct and mm-ReAct do not necessarily break down a given MMH question into a chain of easy to hard unimodal single-hop (USH) sub-questions which can further results in minimum number of calls to the tools.

% One possible reason that LLMs are not well capable to handle MMH reasonings is their limitations to access required information. Providing a set of predefined tools and making them available to LLMs is considered as another plausible solution to enhance multi-hop reasoning ability in LLMs. This family of solutions are simply started by adding information retrieval and web search abilities as accessible tools for LLMs \cite{xu2023search, pereira2023visconde, peng2023check} and can go further by adding more tools  to LLMs \cite{paranjape2023art, schick2023toolformer, zhang2023evaluating, wu2023visual}. Although adding tools empowers LLMs to access more information, it is not optimized for a complex interactive tools-based scenario due to the complexities of reasoning, modality variations, and the accumulation of errors \cite{lu2023chameleon,yao2022react}. To improve the efficiency and effectiveness MMH question-answering model, it is essential to make a chain of dynamic reasonings while interacting with tools and combining their outputs to efficiently and effectively find final answers. 

A possible explanation for the poor performance of LLMs on MMH reasoning tasks is their limitation to access the relevant information. A potential solution to this problem is to equip LLMs with a set of predefined tools that can help target and process specialized information. This approach can include adding information retrieval and web search capabilities to LLMs \cite{xu2023search, pereira2023visconde, peng2023check} or providing LLMs with more advanced tools \cite{paranjape2023art, schick2023toolformer, zhang2023evaluating, wu2023visual}. However, adding tools to LLMs does not guarantee a successful MMH question-answering (QA) solution, because there are still challenges related to complex reasoning, modality variations, and the accumulation of errors \cite{lu2023chameleon,yao2022react}.
To improve the MMH QA model, it is essential to enable LLMs to perform a dynamic reasoning chain that interacts with tools and integrates their outputs to generate the final answers efficiently and effectively. Another important factor contributing in the success of LLMs in reasoning tasks is the model size, where the higher model's capacity leads to a more powerful reasoning ability \cite{stolfo2022causal,shridhar2022distilling,magister2022teaching}.

\begin{figure*}
    \begin{center}
      \includegraphics[width=16cm,height=8cm]{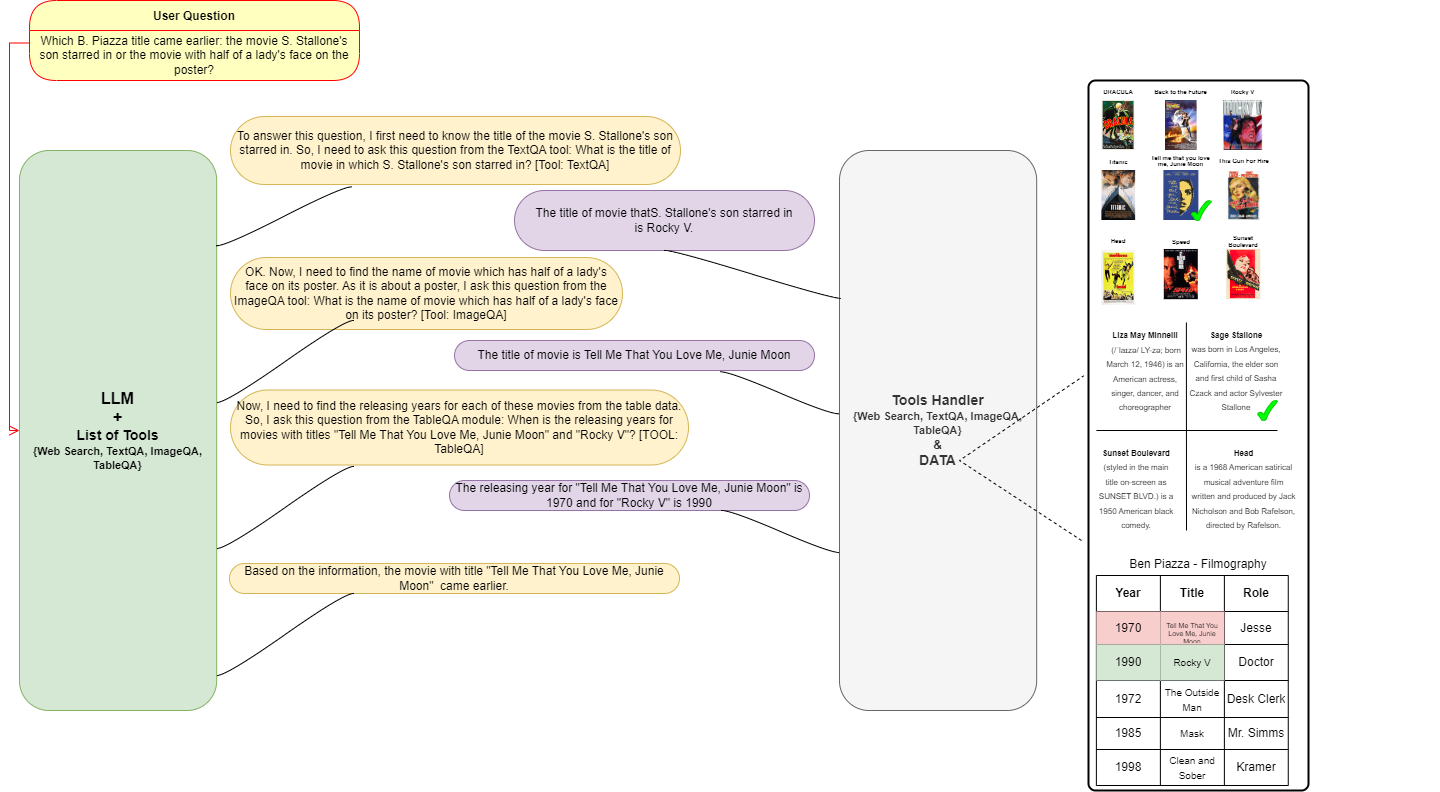}
      \caption{An illustration of the interactive strategy  between LLMs and Tools to answer a MMH question. The LLM first divides the question into a USM sub-question and determines its associated tool. The sub-question is then answered by the associated tool, replying the answer back to LLM. The LLM then asks its next USH question based on the tool's reply.}
    \label{workflow}
    \end{center}
\end{figure*}

In this work, we mainly focus on the problem of MMH question answering using an interactive strategy between LLMs and a set of tools. Figure \ref{workflow} illustrates our proposed strategy with an example.  First, the LLM receives the input MMH question and simplifies it into a unimodal single-hop (USH) sub-question while specifying the tool's name required to answer that sub-question. A tools handler\footnote{We simply postprocess the LLM's output to extract the sub-question and its associated tool.} then receives this USH sub-question, calls the associated tool, and returns the answer. The LLM uses the answer to generate the next USH sub-question. The interaction between the LLM and tools continues until the final answer is found. 

We evaluate this strategy on two complex MMH datasets and benchmark the performance of three LLMs on these datasets. Therefore, the primary contributions of this study can be summarized as follows:
\begin{itemize}
    \item We propose an interactive strategy that enables LLMs to communicate with tools and generate a sequence of sub-questions through a divide-and-conquer approach, allowing LLMs to decompose the MMH questions into USH sub-questions and answer the original question.
    \item We generate a divide-and-conquer dataset to efficiently finetune smaller-size LLMs, thereby enhancing their reasoning capabilities on MMH QA tasks.
    \item We evaluate our strategy on two recent MMH QA datasets and compare the results using LLMs of different sizes.
\end{itemize}

\section{Tool-interacting Divide-and-Conquer Prompting for MMH QA}

Let us consider a scenario where a question requires the use of multiple tools to be answered, i.e. multimodal reasoning. We also assume that the order of tool invocation matters, such that the output of tool $A$ may serve as the input for tool $B$, exemplifying multi-hop reasoning. Furthermore, we define tools as USH QA models that can answer a USH question based on the provided data.

To answer MMH questions, LLMs need to extract an initial USH sub-question that can be addressed with the corresponding tool. The feature of unimodality ensures the LLM calls the correct tool for a sub-question. Moreover, the simplicity feature increases the chance that the LLM receives the correct answer from the tools. Ultimately, the interactive fusion of these two attributes provides a divide-and-conquer strategy, successfully should lead the LLM to the final answer. 

\subsection{LLM as the Divider}
Assuming the example in Figure \ref{workflow}, the LLM first divides the MMH original question into a USH sub-question by asking for the movie's title from the relevant tools. After receiving the tool's reply, it asks the next sub-question, forming another piece of information required for building the final answer. The LLM continues such division behaviour till it gets all necessary pieces of information to answer the original question. 
\subsection{Tools as the Conqueror}
Considering the example in Figure \ref{workflow}, each time that the LLM asks a sub-question, the corresponding tool is invoked to find the requested answer. As the sub-question is a USH question, it is highly likely that the tool successfully obtains the answer. This behaviour introduces tools as a powerful conqueror for the divider. Here, we assume that tools have access only to their associated data modalities, and the answer is given in the corresponding modality.

\section{Efficiently finetuning LLMs}
To enhance the reasoning and tool-interacting capabilities of typical-sized LLMs, such as 7, 13, 30, and 40 billions \footnote{For simplicity, we denote various Language Model sizes as 7b, 13b, 30b, and 40b.}, we efficiently finetune LLMs of different sizes for one epoch using QLoRA \cite{dettmers2023qlora} on a tool-interacting divide-and-conquer dataset ,explained in Section \ref{dataset}. This one epoch of finetuning encourage the corresponding LLM to follow the divide-and-conquer strategy while interacting with the required tools.

\section{Generating a Tool-interacting Divide-and-Conquer Dataset}
\label{dataset}
To build a dataset that answers MMH questions through the proposed tool-interacting divide-and-conquer strategy, we prompted ChatGPT by providing it with manually created few-shot examples. We use eight-shot examples, consisting of three main modality of Text, Table, and image data. Additionally, four different tools, described in Subsection \ref{Tools}, are considered to interact with the associated LLM. 
Table \ref{one-shot-example} presents one of such shots used in prompting ChatGPT. Each shot consists of several steps. Different colors are used to increase the readability of each step. Additionally, the same instruction is used in each shot (illustrated by the yellow color). The main MMH question is depicted in the purple row. The green rows form the division steps (dividing the main MMH question into a USH question), and the red rows represent the conquer steps (the interaction between the LLM and tools). 

For each benchmark reported in Table \ref{tab:accents}, we prompted ChatGPT and pass a random subset\footnote{The size of training subset is 2k for each benchmark.} of the corresponding training set and proceed to generate a tool-interacting divide-and-conquer dataset. 
\begin{table}[h]
\label{one-shot-example}
\begin{center}
\resizebox{\columnwidth}{!}{%
\begin{tabular}{|c|c|}\hline
\textbf{Steps} & \textbf{Description} \\\hline
Instruction & \cellcolor{yellow!15} \begin{tabular}{@{}c@{}} Your task is to answer a given Open Question. However, the Open Question needs \\ \cellcolor{yellow!15} information from different modules, which are Image module, Text module, and Text module.\\ \cellcolor{yellow!15} You can divide the Open Question into a simpler Sub-question and wait to receive the answer it.\\ \cellcolor{yellow!15} Each time you can ask a Sub-question from only one module. \\ \cellcolor{yellow!15}If the provided answer was not complete, you ask a new Sub-question from different or the same module. \\ \cellcolor{yellow!15}You can repeat this loop until you get the complete information to answer the Open Question. 
\end{tabular}
 \\\hline
Question & \cellcolor{blue!15} \begin{tabular}{@{}c@{}} Open Question: 
Which opposing NFL teams that played the Dallas \\ \cellcolor{blue!15} Cowboys at Cowboys Stadium in the 2012 regular season also did not have cheerleaders in 2017?
\end{tabular}
\\\hline
LLM (Divider) & \cellcolor{green!15} \begin{tabular}{@{}c@{}} To answer this question, I first need to know the name of teams that did not have cheerleaders in 2017 from the text module. \\  \cellcolor{green!15} So, I need to ask this sub-question: Sub-question: Which teams did not have cheerleaders in 2017? (Tool=Text) 
\end{tabular}
\\\hline
Tool (Conqueror) & \cellcolor{red!15} \begin{tabular}{@{}c@{}} Answer from the Text Tool: 
Buffalo Bills, Chicago Bears, Cleveland Browns, Green Bay Packers, \\ New York Giants, \cellcolor{red!15} and the Pittsburgh Steelers.
\end{tabular}
\\\hline
LLM (Divider) & \cellcolor{green!15} \begin{tabular}{@{}c@{}} Now, I need to know the name of teams that played against Dallas Cowboys at Cowboys Stadium in the 2012 regular season. \\ \cellcolor{green!15} So, I need to ask this sub-question: Which teams played against Dallas Cowboys at Cowboys Stadium? (Tool=Text)
\end{tabular}
\\\hline
Tool (Conqueror) & \cellcolor{red!15} Answer from the Text Tool: 
Chicago Bears, Cleveland Browns, Persepolis.\\\hline
 LLM (Answer) & \cellcolor{green!15} Based on the answers, the final answer is: [Answer: Chicago Bears, Cleveland Browns]\\\hline
\end{tabular}
}
\caption{A sample shot used in few-shots prompting ChatGPT for generating a tool-interacting divide-and-conquer dataset.}
\end{center}
\end{table}

\section{Experiments}
This section evaluates the effectiveness of our tool-interacting divide-and-conquer strategy for several large language models on two recent MMH QA benchmarks. The comparative analysis involves different combinations of Language Models with varying sizes and strategies. The results are compared in terms of exact matching (EM), $F-$scores ($F_1$), and average number of Tool Calls, i.e. the average number of times that the LLM calls a tool to answer a question. The subsequent sections provide more details about the benchmarks, LLMs, and the obtained results.

\subsection{MMH QA Benchmarks}
We employ two MMH QA benchmarks for evaluation and comparison: MultiModalQA \cite{talmor2021multimodalqa} and MMCoQA \cite{li2022mmcoqa}. These benchmarks provide different data modalities while offering complex multi-hop questions.  Specifically, MultiModalQA contains 29,918 question-answer pairs and encompasses three distinct modalities, namely text data, table data, and image data. Notably, each question in this dataset requires the integration of varying combinations of text, table, and image inputs for accurate answering. Furthermore, MMCoQA is a multimodal conversational QA benchmark, incorporating four modalities: text, table, image, and conversation. This benchmark comprises 1,179 conversations, with an average of 4.88 question-answer pairs per conversation. Table \ref{tab:accents} summarizes these benchmarks.
\begin{table}
\centering
\begin{tabular}{lccc}

\hline
\textbf{Dataset} & \textbf{Modality} & \textbf{Train}  & \textbf{Dev;Test}\\
\hline
MultiModalQA & Text/Table/Image & {23,817} & {2441;3660} \\
MMCoQA & {Conv/Text/Table/Image} & {4582} & {581;590}\\
\hline
\end{tabular}
\caption{Benchmarks details for two challenging MMH QA tasks.}
\label{tab:accents}
\end{table}

\subsection{LLMs}
To assess the effectiveness of our tool-interacting divide-and-conquer strategy, we employ five distinct LLMs for evaluation purposes: StableLM-7b\footnote{\url{https://huggingface.co/OpenAssistant/stablelm-7b-sft-v7-epoch-3}}, Pathia-12b\footnote{\url{https://huggingface.co/OpenAssistant/oasst-sft-4-pythia-12b-epoch-3.5}}, LLaMA-13b \footnote{\url{https://huggingface.co/decapoda-research/llama-13b-hf}}, Falcon-40b \footnote{\url{https://huggingface.co/OpenAssistant/falcon-40b-sft-top1-560}}, and ChatGPT. Except for ChatGPT, other LLMs are finetuned for one epoch on the associated tool-interacting divide-and-conquer datasets \footnote{In the case of mm-ReAct, the tool-interacting dataset is generated in accordance to mm-ReAct strategy}. To assess the reasoning capability of the proposed strategy, we compare it with two other strategies. In the first strategy, the original question is independently processed by each tool, and the returned answers are considered as a prompt to the LLM. Subsequently, the LLM is asked to answer the original question given the prompt. We call this setting \textit{ToolsAnswer}. The second strategy is mm-ReAct, where we follow the vanilla mm-ReAct presented with the tools descriptions. Additionally, we apply our proposed strategy to each LLM, denoted as \textit{Ours} for reference. 
% Due to the inherent limitations of the in-context learning capacity exhibited by StableLM-7b and Pathia-12b, we fine-tune   \textit{Ours}+StableLM-7b and \textit{Ours}+Pathia-12b for one epoch over a small set of data generated by ChatGPT\footnote{We use 3k data generated by \textit{Ours}$+$ChatGPT method to fine-tune StableLM-7b and Pathia-12b.}. This fine-tuning step enables them to split a question into sub-questions and identify the corresponding modalities. Finally, we use few-shot\footnote{We prompted the LLM with eight shots for each dataset.} learning for \textit{Ours}+ChatGPT.

\subsection{Tools}
\label{Tools}
Each data modality needs a specific tool to handle that particular modality. As there are three main modalities, we employ three generic tools\footnote{In the case of baseline methods, we consider the same tools reported in their papers.}, supplemented by an additional Web Search\footnote{Google Search API. \url{https://serpapi.com/}} tool:
\begin{itemize}
    \item \textbf{TextQA} uses Instructor-large\footnote{\url{https://huggingface.co/hkunlp/instructor-large}} \cite{su2022one} which is a text embedding model that has undergone fine-tuning specifically for instructional purposes. Instructor-large produces task-specific and domain-sensitive embeddings by considering a textual input and its corresponding task instructions. We use TextQA for answering a given question based on the text modality.
    
    \item \textbf{TableQA} employs TAPAS\footnote{\url{https://huggingface.co/google/tapas-base-finetuned-wtq}} \cite{herzig2020tapas}, utilizing BERT's structure for covering table-based QA. We employ TableQA as a dedicated tool to address queries associated with tabular data.
    \item \textbf{ImageQA} is based on BLIP-2\footnote{\url{https://huggingface.co/docs/transformers/model_doc/blip-2}} \cite{li2023blip}. This tool receives an image and a question as inputs and returns the corresponding text answer as the output.
    \item \textbf{Web Search} tool is invoked when the remaining tools fail to provide informative answers. In such case, LLM is allowed to do a web search request, retrieving relevant information from the internet.
\end{itemize}

\noindent It is noteworthy to emphasize that these tools are exclusively selected to answer USH questions, ensuring the establishment of a robust basis for comparative evaluations when scrutinizing the effectiveness of our strategy. Moreover, they are not fine-tuned (we consider their original checkpoints). Furthermore, a predefined upper limit is imposed on the number of tool calls to restrict the LLM from exceeding this threshold.  More accurately, the total number of tool calls that the LLM is allowed to answer a given question is limited by this threshold.

\subsection{Results}
This subsection presents the results and compares them in terms of efficacy and accuracy. The evaluation of different methods over MultiModalQA and MMCoQA datasets are reported in Table \ref{tab:T1} and Table \ref{tab:T2}, respectively. In particular, our strategy (labeled as "Ours") consistently outperforms other strategies across LLMs of different sizes, as indicated by higher Exact Match (EM) scores and $F_1$ scores. Additionally, the "Average Tool Calls" column demonstrates that our strategy maintains a relatively low number of tool calls, indicating efficiency in resource utilization \footnote{In case of ToolsAnswer, the Average Tool Calls is always equal to 4, as this strategy calls each tool only once.}.

\begin{table}
\centering
%\resizebox{\textwidth/2}{!}{
\begin{tabular}{lccccc}
\hline
\textbf{LLM} & \textbf{Size} &\textbf{Strategy} & \textbf{EM} & \textbf{$F_1$} &Average Tool Calls\\
\hline
\hline
StableLM &7b &\textit{ToolsAnswer} & {0.0} & {9.36} &{4}\\
Pathia &12b &\textit{ToolsAnswer}  & {0.3} & {15.24} &{4}\\
LLaMA &13b &\textit{ToolsAnswer}  & {1.21} & {17.32} &{4}\\
Falcon &40b &\textit{ToolsAnswer}  & {14.12} & {33.92} &{4}\\
ChatGPT &NA &\textit{ToolsAnswer} &{15.45} &{46.42} &{4}\\
\hline
StableLM$^*$ &7b &\textit{mm-ReAct} & {1.32} & {14.50} &{11.52}\\
Pathia$^*$ &12b &\textit{mm-ReAct}  & {6.51} & {23.18} &{10.60}\\
LLaMA$^*$ &13b &\textit{mm-ReAct}  & {9.87} & {27.45} &{10.53}\\
Falcon$^*$ &40b &\textit{mm-ReAct}  & {18.94} & {45.34} &{8.24}\\
ChatGPT &NA &\textit{mm-ReAct} &{21.30} &{52.16} &{6.22}\\
\hline
StableLM$^*$ &7b &\textit{Ours} & {18.50} & {25.12} &{9.69}\\
Pathia$^*$ &12b &\textit{Ours}  & {21.32} & {31.14} &{8.14}\\
LLaMA$^*$ &13b &\textit{Ours}  & {23.14} & {35.21} &{8.75}\\
Falcon$^*$ &40b &\textit{Ours}  & {41.18} & {56.74} &{6.34}\\
ChatGPT &NA &\textit{Ours} &{43.71} &{61.03} &{5.07} \\
\hline
Human &{-} &{-} &{86.2} &{91.2} &{-}\\
\hline
\end{tabular}
%}
\caption{Validation results on the MultiModalQA benchmark. All methods are presented with an identical set of tools. "NA" denotes data that is not available. The maximum number of Tool Calls per each question is set to 12. "*" means the corresponding LLM is finetuned using QLoRA for one epoch through its associated strategy.}
\label{tab:T1}
\end{table}

\begin{table}
\centering
%\resizebox{\textwidth/2}{!}{
\begin{tabular}{lccccc}
\hline
\textbf{LLM} & \textbf{Size} &\textbf{Strategy} & \textbf{EM} & \textbf{$F_1$} &Average Tool Calls\\
\hline
\hline
StableLM &7b &\textit{ToolsAnswer} & {0.0} & {8.35} &{4}\\
Pathia &12b &\textit{ToolsAnswer}  & {0.0} & {12.41} &{4}\\
LLaMA &13b &\textit{ToolsAnswer}  & {0.0} & {13.95} &{4}\\
Falcon &40b &\textit{ToolsAnswer}  & {3.45} & {22.15} &{4}\\
ChatGPT &NA &\textit{ToolsAnswer} &{8.91} &{46.10} &{4}\\
\hline
StableLM$^*$ &7b &\textit{mm-ReAct} & {2.24} & {17.42} &{11.56}\\
Pathia$^*$ &12b &\textit{mm-ReAct}  & {6.15} & {18.51} &{10.35}\\
LLaMA$^*$ &13b &\textit{mm-ReAct}  & {7.84} & {21.32} &{10.41}\\
Falcon$^*$ &40b &\textit{mm-ReAct}  & {18.37} & {41.08} &{8.96}\\
ChatGPT &NA &\textit{mm-ReAct} &{41.33} &{52.17} &{6.35}\\
\hline
StableLM$^*$ &7b &\textit{Ours} & {7.11} & {16.36} &{10.22}\\
Pathia$^*$ &12b &\textit{Ours}  & {10.76} & {24.21} &{9.34}\\
LLaMA$^*$ &13b &\textit{Ours}  & {11.27} & {26.39} &{6.54}\\
Falcon$^*$ &40b &\textit{Ours}  & {38.91} & {56.51} &{4.96}\\
ChatGPT &NA &\textit{Ours} &{47.05} &{58.82} &{3.80} \\
\hline
Human &{-} &{-} &{NA} &{NA} &{-}\\
\hline
\end{tabular}
%}
\caption{Test results on the MMCoQA benchmark. All methods are presented with an identical set of tools. "NA" denotes data that is not available. The maximum number of Tool Calls per each question is set to 12. "*" means the corresponding LLM is finetuned using QLoRA for one epoch through its associated strategy.}
\label{tab:T2}
\end{table}

% \subsection{Ablation Study}
% Table \ref{tab:T1} reports a competitive performance for the baseline method, i.e. ImplicitDecomp. To further investigate this observation, a comparison between ImplicitDecomp and our strategy is conducted on a small, randomly sampled subset from the MMCoQA dataset. The outcomes are depicted in Table \ref{tab:ablation}, revealing that the performance of our strategy is superior over ImplicitDecomp. We argue that the competitive performance achieved on MultiModalQA by ImplicitDecomp is attributed to training of both the method and its tools on the provided training data. Consequently, it cannot generalize well on another dataset, while our strategy is not limited with training, thus enabling enhanced generalization capabilities.

% \begin{table}
% \centering
% %\resizebox{\columnwidth}{!}{
% \begin{tabular}{lccc}
% \hline
% \textbf{Method} & {EM} & {$F_1$}\\
% \hline
% ImplicitDecomp (baseline) & {16.18} &{22.34}\\
% \hline
% \textit{Ours}+ChatGPT  & {44.11} & {58.25}\\
% \hline
% \end{tabular}
% %}
% \caption{Comparing the obtained results on a randomly selected subset of 50 instances extracted from the MMCoQA dataset. ImplicitDecomp uses its own pre-trained weights. This observation shows that the performance of our strategy does not dependent on pre-training steps.}
% \label{tab:ablation}
% \end{table}

\section{Conclusion}
This study presents a tool-interacting  strategy, leveraging a divide-and-conquer interaction between large language models and a set of tools to effectively answer multimodal multi-hop questions. Our strategy facilitates the division of MMH questions into USH sub-questions, allowing LLMs to interact with a predefined set of tools for obtaining intermediate answers. We assessed the performance of different-sized LLMs in three different reasoning strategies. The obtained results demonstrate the effectiveness of our strategy. For the possible future directions, we will include exploring inter-tool communication, handling non-predefined modalities, and improving the performance of smaller LLMs for MMH QA tasks.

\bibliographystyle{unsrt}  
%\bibliography{references}  %%% Remove comment to use the external .bib file (using bibtex).
%%% and comment out the ``thebibliography'' section.

%%% Comment out this section when you \bibliography{references} is enabled.
\bibliography{references}

\end{document}